# SHRAV: State-Hypothesis-Reason-Action-Verify Framework for Physical Modeling and Inverse Design

Ziheng Guo[1*], Yang Bu[1*]
[*]ZhangJiang Laboratory

## Abstract

Physical modeling and inverse design require computation that can continue from reusable state. We introduce **SHRAV**, an architecture-independent computational framework organized around State, Hypothesis, Reason, Action, and Verify. Its central mechanism is a **state-continuation core** with declared reuse boundaries and explicit roles for learned evolution and numerical quantities. Forward configurations evolve predictive state and read out physical responses; inverse-design configurations additionally generate target-directed modifications and consume evaluator feedback. Electromagnetic world-model studies are mapped to forward configurations, with selected readout and reuse diagnostics reported here. Computational lithography demonstrates an inverse-design configuration: four fixed-weight design updates improve thresholded aerial-image intersection-over-union from 0.5313 to 0.8153 under independent scalar-pupil replay, with a maximum absolute IoU difference of approximately 0.000824 between predictor estimates and independent replay.

## 1. From physical prediction to design evolution

Inverse design seeks a structure whose response meets a target under stated conditions and constraints [1,2]. SHRAV treats a continuing physical computation, including design, as the computational object: what persists, what is computed internally, what changes in the physical design, and how its evaluated consequences enter the next step.

**The proposed separation concerns representation and computational responsibility.** Learned latent evolution need not reproduce the complete numerical solution. Numerical variables and observations can retain their own meaning and enter continuation through an explicit interface. In inverse-design configurations, the evolved internal representation, executed modification, physical observation, and target-relative feedback jointly inform the next state. This separation does not assert statistical independence between latent and physical quantities or guarantee transfer across conditions.

Our earlier electromagnetic world-model studies [3,4] provide the research context. Under the configuration taxonomy defined here, **FieldSeer I** [3] maps to a forward configuration with externally supplied geometry interventions. Its multi-structure offline waveguide evaluation reports physical-domain field MSE of 0.01122, with 180 observed frames followed by 100 predicted frames [3].

SHRAV is intended to support **AI-native inverse design** through reusable learned computation and evaluator-informed continuation. Computational lithography offers explicit mask actions, spatial discrepancies, and image/contour metrics for demonstrating the interface. The framework itself is not specific to lithography.

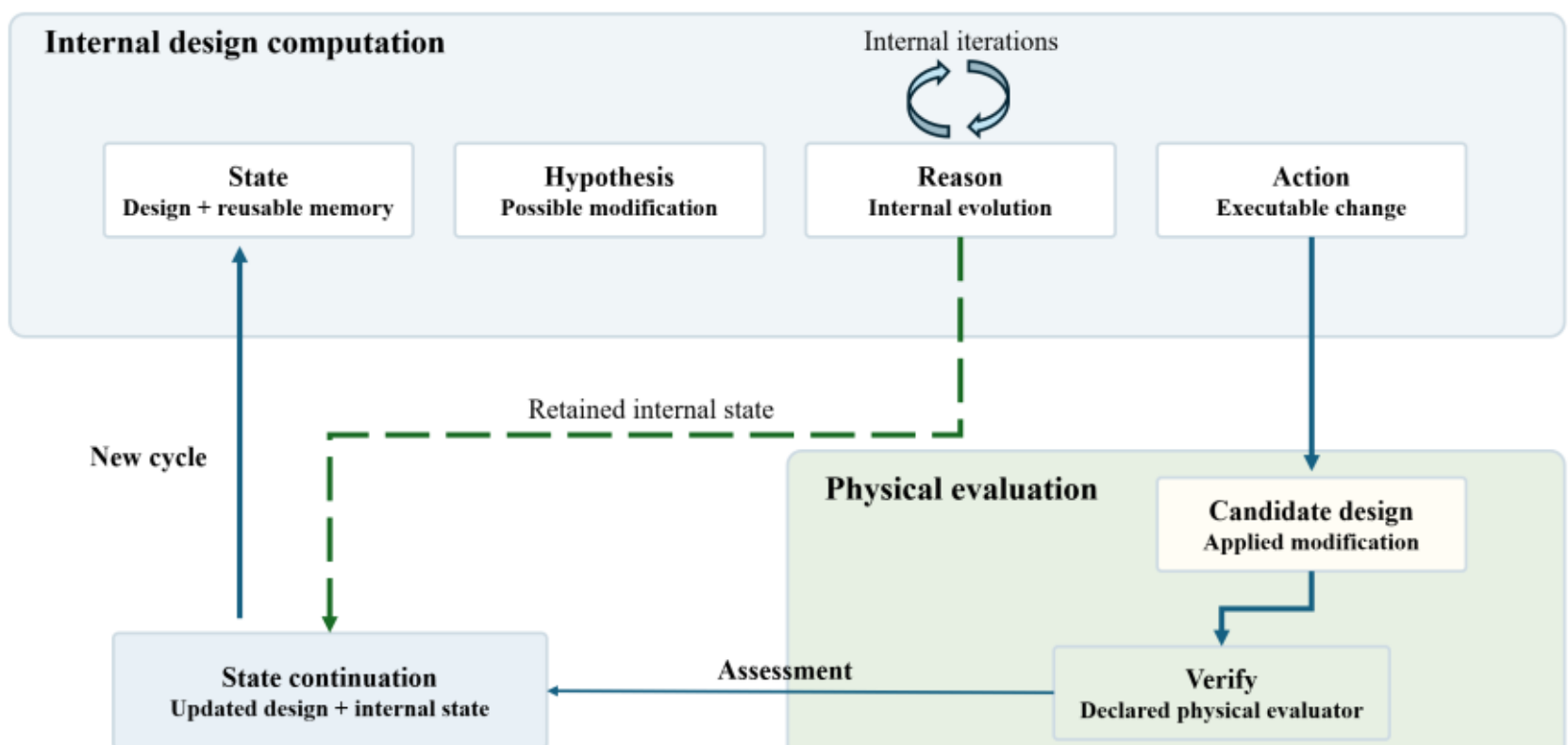

**Figure 1. Shared continuation interface instantiated for inverse design.** Retained internal state from Reason (green dashed path) and evaluator assessment from Verify (blue solid path) enter the state-continuation update $U$. The updated design and internal state initialize the next cycle. **Feedback from update $t$ is an input to update $t+1$; an explicit acceptance gate is not required.** Action produces an executable candidate design, and Verify assesses its physical response using a predictor, numerical solver, or compatible measurement. The circular arrows above Reason denote internal iterations.

## 2. Shared core and configuration contracts

### 2.1 Shared continuation core

**Shared-core conformance.** A SHRAV configuration declares (i) persistent state with an explicit reuse boundary, including conditions for reinitialization, and (ii) distinct computational roles for learned evolution and numerical quantities, including whether numerical readouts or observations re-enter the state update. At least one learned component must participate in the evolution or recontextualization of retained state. This core is necessary but not sufficient for inverse-design conformance.

Separation concerns computational roles, not statistical independence. Existing models may satisfy this contract; the framework does not claim exclusivity over these ingredients. A purely numerical solver alone is not a SHRAV configuration. Numerical solvers can participate as evaluators or internal computational components within a configuration that satisfies the shared core.

Write $S_t = (x_t, z_t, o_t, c_t)$ for the current structure, reusable representation, latest observation or readout, and conditions. Explicit numerical variables and provenance may be included. An inverse-design configuration additionally carries a target $y^\star$. The state need not be a complete solver state or a proven sufficient Markov state.

**Forward configurations** continue prediction under declared conditions; geometry interventions may be externally supplied. Observation feedback can also support forward estimation. **Inverse-design configurations** select and execute modifications toward a target and consume their evaluated consequences in subsequent design computation. Feedback alone does not make a predictor an inverse designer.

In a forward configuration, State and Reason retain and evolve the learned representation; Verify provides numerical readout and, where specified, assessment against observations or a reference. Hypothesis may reduce to a prescribed continuation, while Action may be externally supplied or inactive. FieldSeer's reported geometry interventions belong to this externally supplied case. The additional inverse-design criteria apply when the system selects and executes modifications toward a target.

### 2.2 Inverse-design roles and criterion

**State** retains and contextualizes information. **Hypothesis** proposes a correction, candidate, or trajectory, possibly implicitly. **Reason** develops it through internal computation, yielding $r_{t,L}$ at declared depth $L$. **Action** produces $a_t$ and an executable candidate $\tilde{x}_{t+1} = \mathcal{T}(x_t, a_t)$. **Verify** supplies an assessment through a qualified predictor, numerical solver, or compatible measurement:

$$\tilde{o}_{t+1} = \mathcal{V}(\tilde{x}_{t+1}, c_t),$$
$$e_{t+1} = \Phi(\tilde{o}_{t+1}, y^\star),$$
$$S_{t+1} = U\big(S_t, r_{t,L}, a_t, \tilde{x}_{t+1}, \tilde{o}_{t+1}, e_{t+1}\big).$$

Where $\Phi$ extracts spatial discrepancies or scalar criteria. Both the evolved representation and assessment enter continuation. A policy may escalate evaluation to a high-fidelity solver when declared reliability or consistency criteria fail. Such fallback is distinct from design rollback and is not exercised in the displayed trajectory.

**Inverse-design conformance.** In addition to the shared core, a SHRAV inverse-design realization declares (i) persistent state reused across executed design updates, beyond time stepping within a fixed geometry; (ii) internal hypothesis development distinct from evaluation; (iii) an executable target-directed design action; (iv) evaluator feedback consumed by $U$; and (v) the evaluator and assessment criteria. Scores used only for reporting do not satisfy (iv). These additional requirements do not apply to every forward configuration.

SHRAV specifies system-level computational roles and their continuation interfaces. A domain-specific realization instantiates these roles using learned models, numerical routines, optimization tools, and, optionally, human interventions. The roles may be integrated within one model or distributed across interacting components; they do not prescribe separate networks or a training paradigm. Adversarial training, diffusion, reinforcement-learning rewards, and adjoint sensitivities remain optional components.

### 2.3 Implemented inverse feedback

The lithographic loop uses a frozen hybrid predictor and retained latent state. In this exemplar, retained latent state is reused across executed mask updates within a rollout at fixed target and optical conditions. $L = 4$ denotes internal Reason iterations per outer design update, distinct from the four outer updates in Figure 4. No optical evaluation occurs between those internal iterations. For target $T$ and threshold image $P_t$, missing regions $T(1 - P_t)$, excess regions $(1 - T)P_t$, and the response enter recontextualization. All candidates are applied with fixed weights and no acceptance gate. Independent scalar-pupil replay audits the saved masks separately.

## 3. Forward configurations of the interface

Figures 2 and 3 instantiate two forward configurations. Together with the inverse exemplar of Figure 4, they illustrate complementary aspects of the interface. Figure 2 exhibits separation: numerical readout refinement leaves the latent trajectory unchanged. Figure 4 exhibits coupling: evaluator feedback re-enters design continuation through $U$. Together, these results support an interface that permits readout refinement without altering recurrent evolution and feedback-conditioned continuation during target-directed design. In Figure 2, readouts share a rollout under fixed geometry and source conditions. In Figure 3, each model continues under fixed mask and optics. Changed histories or unvalidated conditions require reinitialization; neither example establishes unrestricted state transfer. These selected diagnostics are reported here.

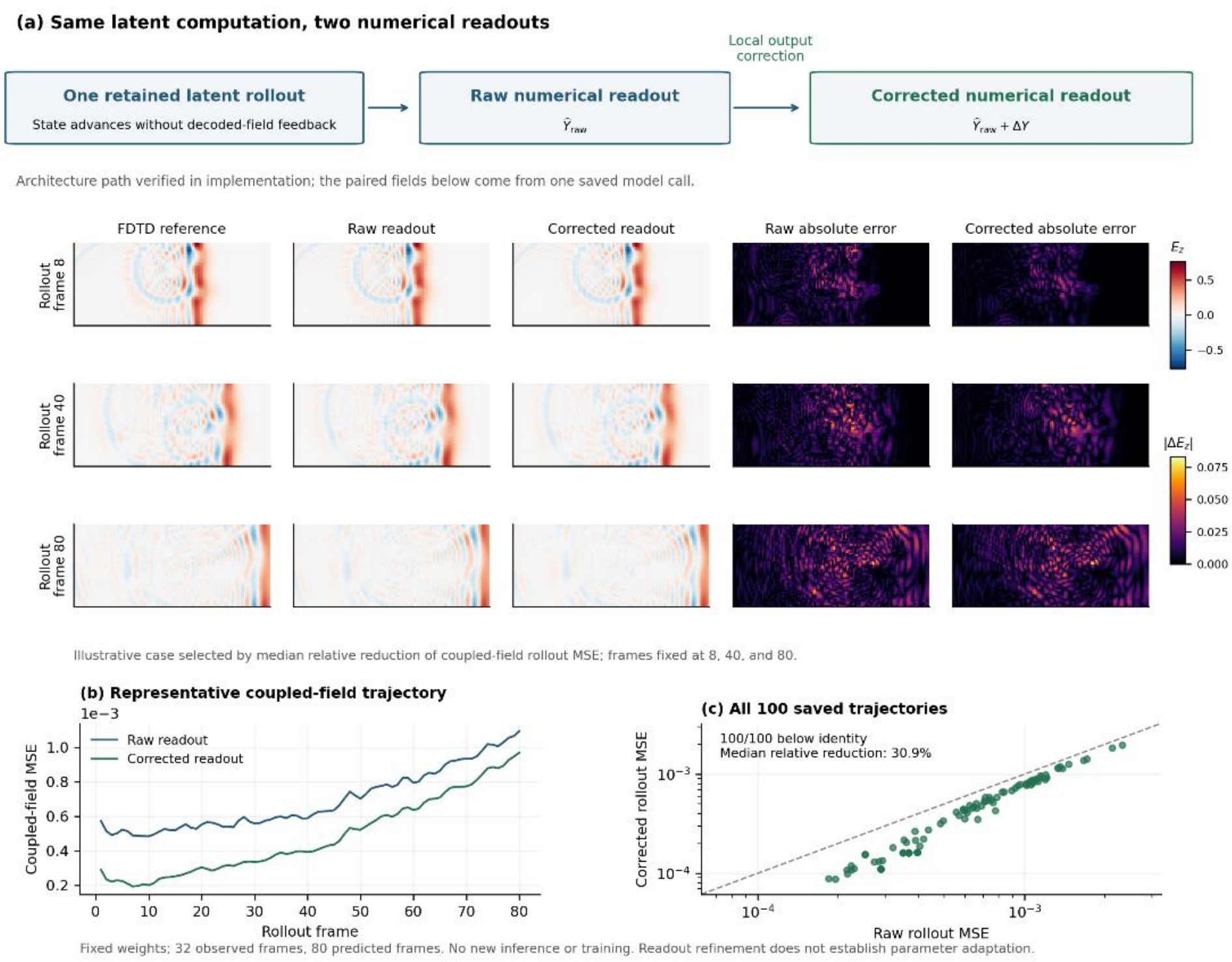

**Figure 2. Waveguide evolution with separate numerical readout refinement.** A coupled-field model assimilates 32 frames and predicts 80. Raw and corrected fields come from the same model call and latent rollout; corrected fields do not feed the recurrent transition. The representative case is near the median: it is the 50th of 100 saved trajectories ranked in ascending order of relative coupled-field MSE reduction. Its $E_z$ fields are shown at rollout frames 8, 40, and 80. Field plots share one symmetric scale; error plots share another. Across all 100 saved trajectories, whole-rollout coupled-field MSE decreases in 100/100 cases, with a median relative reduction of **30.9%**. This paired diagnostic supports readout refinement without changing the latent trajectory.

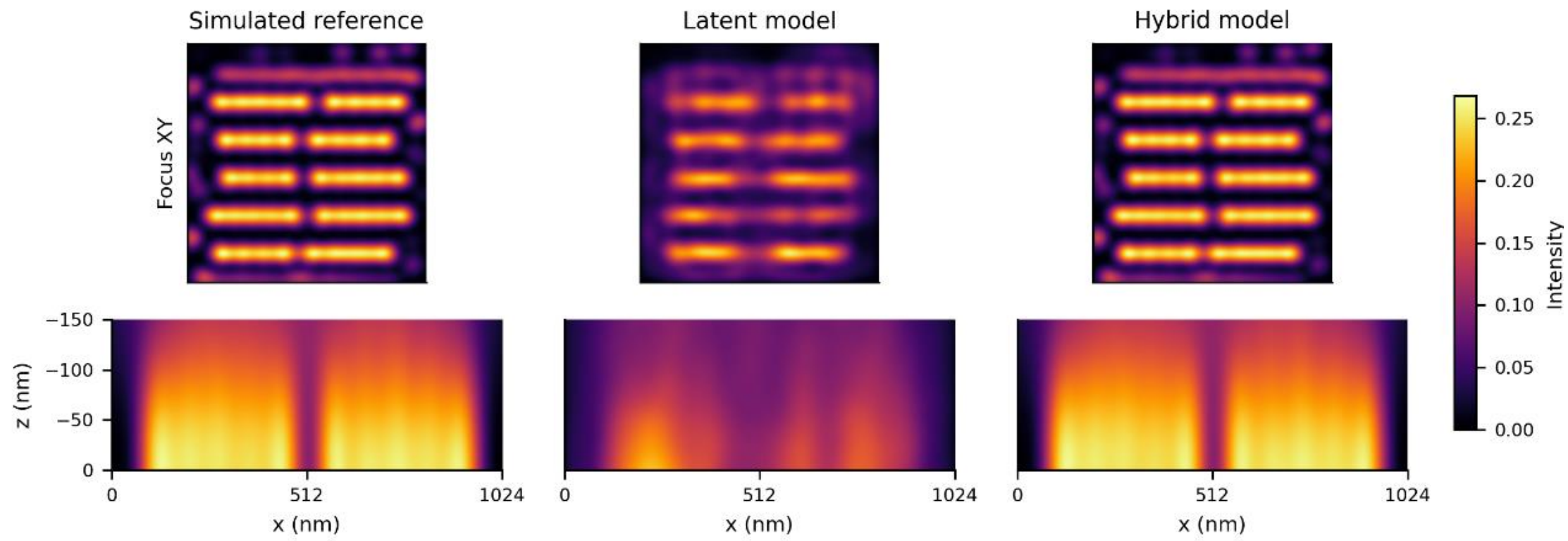

**Figure 3. Lithographic propagation with latent and hybrid models.** Columns show the simulated reference, Latent model, and Hybrid model for the same mask. Rows show focal-plane intensity and the same $x$-$z$ section, with a common intensity scale. The Hybrid model carries an explicit numerical propagation state, providing another way to couple learned computation and numerical quantities. These are saved outputs from separately trained implementations under identical display scales; they illustrate distinct latent-numerical organizations rather than isolate a single cause.

## 4. Numerical exemplar: computational lithography

Figure 4 presents a selected validation trajectory. The intensity threshold **0.225** is held fixed and used consistently in both feedback construction and numerical assessment. Thresholded-image IoU is the intersection divided by the union of predicted and target foreground pixels; EPE P95 summarizes the independently replayed full-contour error in nanometers.

(a) Executed designs and feedback-conditioned continuation

Missing target region | Excess region

Target | Initial design | Update 1 | Update 2 | Update 3 | Update 4

Predicted intensity

Predicted threshold image: IoU 0.5308, IoU 0.6300, IoU 0.7263, IoU 0.7923, IoU 0.8152

Feedback discrepancies: Used for update 1, Used for update 2, Used for update 3, Used for update 4, Terminal assessment

Relative intensity: 0.0, 0.2, 0.4, 0.6

(b) Independent overlap verification

Max. absolute difference: $8.24 \times 10^{-4}$ 0.8153

IoU; Outer design update; Hybrid evaluator; Independent pupil

(c) Independent contour verification

Lowest EPE at update 2

EPE P95 (nm); Outer design update; 21.34, 13.93, 9.53, 11.96, 15.00

**Figure 4. Executed designs, feedback, and numerical verification. (a)** Target, binary masks, predicted focal intensities, thresholded images, and feedback discrepancies. Cyan marks the target; amber indicates missing regions and magenta excess regions. Feedback in columns 0-3 is consumed before the following update; column 4 shows terminal discrepancies. **(b)** Predicted and independent scalar-pupil IoU. **(c)** Independent full-contour EPE P95. Indices denote outer design updates. The figure is reconstructed from the recorded design trajectory using saved responses and the implemented threshold rule. Independent replay shares the nominal optical assumptions and is not a wafer measurement.

Independent IoU follows **0.5313, 0.6305, 0.7271, 0.7932, 0.8153**, giving an absolute gain of **0.2840**. The maximum absolute prediction-replay difference is **0.000824** (rounded). EPE P95 follows **21.34, 13.93, 9.53, 11.96, 15.00 nm**. Overlap favors update 4, whereas contour accuracy favors update 2. These metric-dependent preferences do not imply an executed rollback. The exemplar demonstrates retained-state continuation across executed mask updates, with predictive feedback informing subsequent computation and independent numerical replay assessing the resulting designs.

## 5. Positioning, domains, and progression

Learned optimizers [5], latent world models [6,8], and goal-directed planners [7] provide related ingredients. SHRAV makes the reuse boundary, latent/numerical roles, and their continuation interfaces explicit. Adjoint methods [1,2] can supply sensitivities or local refinement. This is a configuration contract, not a claim that existing model families cannot retain state or combine feedback sources.

### 5.1 Domain mappings and evidence

**Evidence index.** FieldSeer I [3] maps to a forward configuration with externally supplied geometry interventions. Figure 2 reports a companion-developed waveguide readout diagnostic. Figures 3-4 show forward propagation and inverse design. A separate archived companion check compares cached and uncached execution in 30 edited cases. Reusing the prefix encoding and base rollout preserves the compared outputs in all 30 cases, with a recorded maximum absolute difference of 0. This demonstrates execution equivalence, not an independent physical-accuracy test.

**Table 1. Configuration mappings and evidence status.** FDTD and FEM denote numerical methods used within configurations, not SHRAV configurations by themselves. Illustrative inverse mappings do not establish validated implementations.

| Task / evidence | Configuration | Action and evaluation interface | Status |
|---|---|---|---|
| FieldSeer I [3,4] | Forward, geometry editable | Supplied geometry edit; field forecast | Retrospective mapping of public results |
| Waveguide readout, Figure 2 | Forward | Fixed-condition latent rollout; numerical readout correction | Paired diagnostic reported here |
| Lithographic propagation, Figure 3 | Forward | Fixed mask/optics; latent or hybrid propagation | Saved examples reported here |
| Mask refinement, Figure 4 | Inverse, Level-0 | Target-directed mask update; predictive feedback and separate pupil audit | Design loop demonstrated here |
| Waveguide inverse design / inverse scattering | Inverse | Geometry/material update; FDTD/FEM or measurement-compatible assessment | Illustrative mappings |

**Companion progression.** Two companion directions are role-attribution studies across physical domains and experimental-feedback adaptation under shifted process conditions. Subsequent publications will present mechanisms, evaluation protocols, and limitations.

### 5.2 Capability levels and scope

Configuration and adaptation level are orthogonal: Level-0 establishes fixed-weight state continuation; Level-1 additionally establishes reuse of learned computational operators across a task family; Level-2 additionally establishes parameter adaptation assessed under declared criteria before adoption. These are cumulative capability requirements: a fixed-weight rollout may also belong to a Level-1 realization. The displayed inverse loop establishes Level-0 conformance.

**AI-native inverse design, within the SHRAV taxonomy,** denotes target-directed processes in which reusable learned state and evaluator feedback jointly condition continuation.

Level-2 identifies their verified self-adaptive form. An implementation qualifies as a SHRAV Level-2 inverse-design realization when it satisfies the shared-core and inverse-design criteria of Section 2, reuses learned operators across a task family, and verifies parameter updates against declared criteria before adoption. This report defines the base interface and classification; detailed self-adaptation mechanisms and evaluations are reserved for companion studies.

## 6. Conclusion

SHRAV specifies a continuation core with explicit reuse boundaries and latent/numerical roles. Forward configurations expose predictive evolution and readout; inverse-design configurations additionally execute target-directed changes and consume evaluation feedback. Their common interface connects reusable learned computation with physical assessment without binding the framework to one network or domain.